\documentclass{article}

\usepackage{arxiv}

\usepackage[utf8]{inputenc} 
\usepackage[T1]{fontenc}    
\usepackage{hyperref}       
\usepackage{url}            
\usepackage{booktabs}       
\usepackage{amsfonts}       
\usepackage{nicefrac}       
\usepackage{microtype}      
\usepackage{graphicx}
\usepackage{natbib}
\usepackage{doi}
\usepackage{pifont}
\usepackage{float}
\usepackage{bbding}
\usepackage{amsmath}
\usepackage{lipsum}
\graphicspath{ {./figures/} }
\usepackage{authblk}

\title{StoryMaker: Towards Holistic Consistent Characters in Text-to-image Generation}

\date{} 					

\author{Zhengguang Zhou, Jing Li, Huaxia Li\thanks{Project Leader}, Nemo Chen, Xu Tang\\
\small Xiaohongshu Inc.\\
}

\renewcommand{\headeright}{}
\renewcommand{\undertitle}{Technical Report}

\hypersetup{
pdftitle={A template for the arxiv style},
pdfsubject={q-bio.NC, q-bio.QM},
pdfauthor={David S.~Hippocampus, Elias D.~Striatum},
pdfkeywords={First keyword, Second keyword, More},
}

\begin{document}
\maketitle

\begin{figure}[htb]
	\centering
        \includegraphics[scale=0.48]{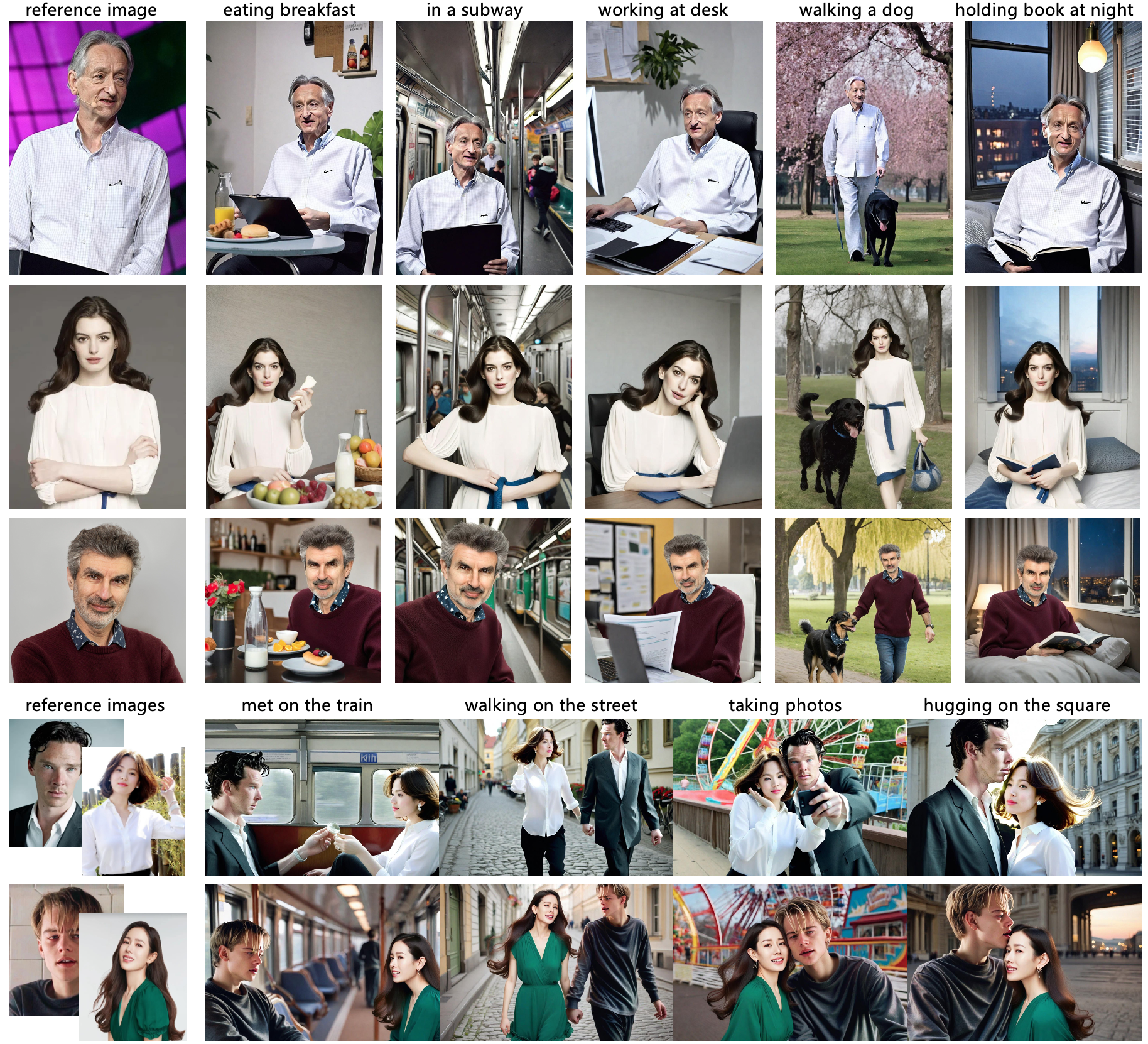}
	\caption{Visualization of images generated by our StoryMaker. The first three rows depict a story about a day in the life of an "office worker," while the last two rows tell a story inspired by the movie "Before Sunrise."}
	\label{fig:day1}
\end{figure}

\begin{abstract}
Tuning-free personalized image generation methods have achieved significant success in maintaining facial consistency, $i.e.$, identities, even with multiple characters. However, the lack of holistic consistency in scenes with multiple characters hampers these methods' ability to create a cohesive narrative. In this paper, we introduce StoryMaker, a personalization solution that preserves not only facial consistency but also clothing, hairstyles, and body consistency, thus facilitating the creation of a story through a series of images. StoryMaker incorporates conditions based on face identities and cropped character images, which include clothing, hairstyles, and bodies. Specifically, we integrate the facial identity information with the cropped character images using the Positional-aware Perceiver Resampler (PPR) to obtain distinct character features. To prevent intermingling of multiple characters and the background, we separately constrain the cross-attention impact regions of different characters and the background using MSE loss with segmentation masks. Additionally, we train the generation network conditioned on poses to promote decoupling from poses. A LoRA is also employed to enhance fidelity and quality. Experiments underscore the effectiveness of our approach. StoryMaker supports numerous applications and is compatible with other societal plug-ins. Our source codes and model weights are available at \href{https://github.com/RedAIGC/StoryMaker}{https://github.com/RedAIGC/StoryMaker}.
\end{abstract}

\section{Introduction}
Diffusion-based image generation methods, such as DALL-E \citep{ramesh2021zeroshot}, Imagen \citep{saharia2022photorealistic}, and Stable Diffusion \citep{rombach2022highresolution}, have recently made significant advancements. However, personalizing generated content using texts alone remains challenging. To address this, test-time fine-tuning methods \citep{avrahami2023break,gal2022image,kumari2023multi,ruiz2023dreambooth} have been proposed to produce images with specific subjects. Nevertheless, their generalization ability is constrained by the limited number of images and the high cost of fine-tuning. Consequently, tuning-free methods \citep{li2024photomaker,ma2024subject,wei2023elite,xiao2023fastcomposer,wei2024mm,kim2024instantfamily,ye2023ip,wang2024instantid,han2024emma} trained on large-scale datasets have been introduced. These methods employ a visual encoder to integrate visual information into the generator without the need for lengthy fine-tuning. While \cite{xiao2023fastcomposer,wei2024mm,kim2024instantfamily} preserve facial identities, they fail to maintain the holistic consistency including consistent clothing, hairstyles, and bodies, thereby limiting their applications.

In this paper, we introduce StoryMaker, which pursues the holistic consistency, not only preserving facial identities but also clothing, hairstyles, and bodies. StoryMaker allows variation in backgrounds, character poses, and styles through text prompts, enabling the generation of a series of images with consistent characters, thereby creating a narrative. StoryMaker also facilitates applications such as clothing swapping and image variation and is compatible with plug-ins like LoRA for stylization.

To preserve clothing, hairstyles, and bodies in addition to faces, StoryMaker conditions the generation on face identities and cropped character images, which include clothing, hairstyles, and bodies. After extracting information from the reference image, we integrate face identities and cropped character images using the Positional-aware Perceiver Resampler (PPR) to derive character features.

As it is more difficult to retain clothing, hair styles and bodies, other than only face identities, StoryMaker regularizes the cross-attention impact region among different characters as well as the background. Unlike MM-Diff \citep{wei2024mm}, which only separates different foregrounds, we include a learnable background embedding to encourage differentiation from the background. An ID loss is introduced to further regularize identities. To decouple generation from the poses of cropped character images, enhancing diversity and utility, we train our model on predicted poses with ControlNet \citep{zhang2023adding}. During inference, ControlNet can be omitted, allowing poses to be guided directly by text prompts. Alternatively, referred poses can be provided to ControlNet. A LoRA is employed to improve fidelity and quality. By combining these elements, StoryMaker generates image series with consistent faces, clothing, hairstyles, and bodies, thereby constructing a coherent story.

In summary, the main contributions of this paper are: i) We address the task of generating a series of images with consistent faces, clothing, hairstyles, and bodies, while allowing variations in backgrounds, poses, and styles via text prompts, enabling narrative creation. ii) To tackle this complex task, we propose StoryMaker, which first extracts information from reference images and refines it using the Positional-aware Perceiver Resampler. To prevent different characters and the background from interleaving each other, we regularize the cross-attention impact region using MSE loss with segmentation masks and train the backbone network conditioned on poses by ControlNet to facilitate decoupling. We also train a LoRA to enhance fidelity and quality. iii) Experiments demonstrate that our proposed StoryMaker achieves excellent performance and has diverse applications in real-world scenarios.

\section{Related Work}
\subsection{Subject-Driven Image Generation}
Subject-driven text-to-image generation has achieved remarkable progress. Current methods in this domain can be categorized based on whether they necessitate test-time fine-tuning for input images. Early approaches \citep{ruiz2023dreambooth,gal2022image,gal2023encoder} require test-time optimization of specific text tokens to represent target concepts using a limited set of subject images. These fine-tuning methods are time-consuming due to the slow optimization process before inference. Recent methods aim to eliminate the need for fine-tuning by integrating additional modules while keeping the primary pre-trained text-to-image models frozen. Subject-Diffusion \citep{ma2024subject} substitutes text tokens describing subjects with the corresponding image embeddings and trains an adapter module to incorporate fine-grained image features. ELITE \citep{wei2023elite} and FastComposer \citep{xiao2023fastcomposer} also map images to text embeddings by training an additional network. Blip-Diffusion \citep{li2024blip} employs the pre-trained multi-modal encoder BLIP-2 \citep{li2023blip} to infuse image information. IP-Adapter \citep{ye2023ip-adapter} separates image and text features in cross-attention, allowing for independent image feature integration. MoA (Mixture-of-Attention) \citep{ostashev2024moa} enhances image quality by segregating subject and context. The SSR-Encoder \citep{zhang2024ssr} is a recent development that integrates segment information into text features through cross-attention, facilitating selective feature extraction.

Identity-preserving human image generation is a prominent area in subject-driven image generation, given its broad real-world applications. Solutions such as FaceStudio \citep{yan2023facestudio}, IP-Adapter-FaceID \citep{ye2023ip-adapter}, FlashFace \citep{zhang2024flashface}, and PhotoMaker \citep{li2024photomaker} utilize ID embeddings derived from Arcface \citep{deng2019arcface} as a condition, which is crucial for maintaining facial fidelity. The leading approach, InstantID \citep{wang2024instantid}, introduces IdentityNet, which uses five facial keypoints to control face structure, achieving optimal face similarity. Beyond single-ID customization, some methods \citep{wei2024mm,he2024uniportrait,jang2024identity,kumari2023multi,avrahami2023break} focus on multi-ID image generation. Some studies \citep{kim2024instantfamily,kong2024omg} use predefined layouts to guide multi-ID image generation, while limits the scalability in real-world scenarios. In contrast, MM-diff \citep{wei2024mm} imposes constrains on the cross-attention maps with different subjects during the training phase, which guarantees the generation of multi-ID images without any predefined input. Recently, UniPortrait \citep{he2024uniportrait} employs an ID routing module to unify multi-ID customization, avoiding identity blending. Our proposed StoryMaker not only preserves faces in image generation, but also ensures consistency in clothing, hairstyle, and bodies. For multi-character generation, we introduce the Positional-aware Perceiver Resampler and attention loss to address multi-character blending.

\subsection{Image Story Generation}
Maintaining consistent content across a series of generated images has numerous real-world applications, such as story visualization and comic creation. StoryDiffusion \citep{zhou2024storydiffusion} employs a consistent self-attention mechanism that adapts information from other images in the batch to ensure character consistency in storytelling sequences.
Unlike StoryDiffusion, which adapts from full images, ConsiStory \citep{tewel2024training} uses a subject-driven shared attention block that only adapts information from masked subjects, with correspondence-based feature injection enhancing subject consistency between images.
DreamStory \citep{he2024dreamstory} leverages a Large Language Model (LLM) for better understanding and guidance in generation. Subjects are generated first, followed by a Multi-Subject Consistent Diffusion model, ensuring subject consistency across images by adapting information from other images in self-attention and from texts in cross-attention, similar to ConsiStory.
OneActor \citep{wang2024oneactor} introduces a cluster-conditioned generation paradigm, achieving controlled, consistent subject generation by tuning an adapter to inject modified prompt embeddings into the fixed U-Net.
Our method focuses on generating images with consistent subjects given references, while these methods work without references. Approaches like StoryDiffusion, ConsiStory, and DreamStory are training-free, yet backgrounds can easily be involved with inaccurate masks.

\begin{figure}
	\centering
        \includegraphics[scale=0.4]{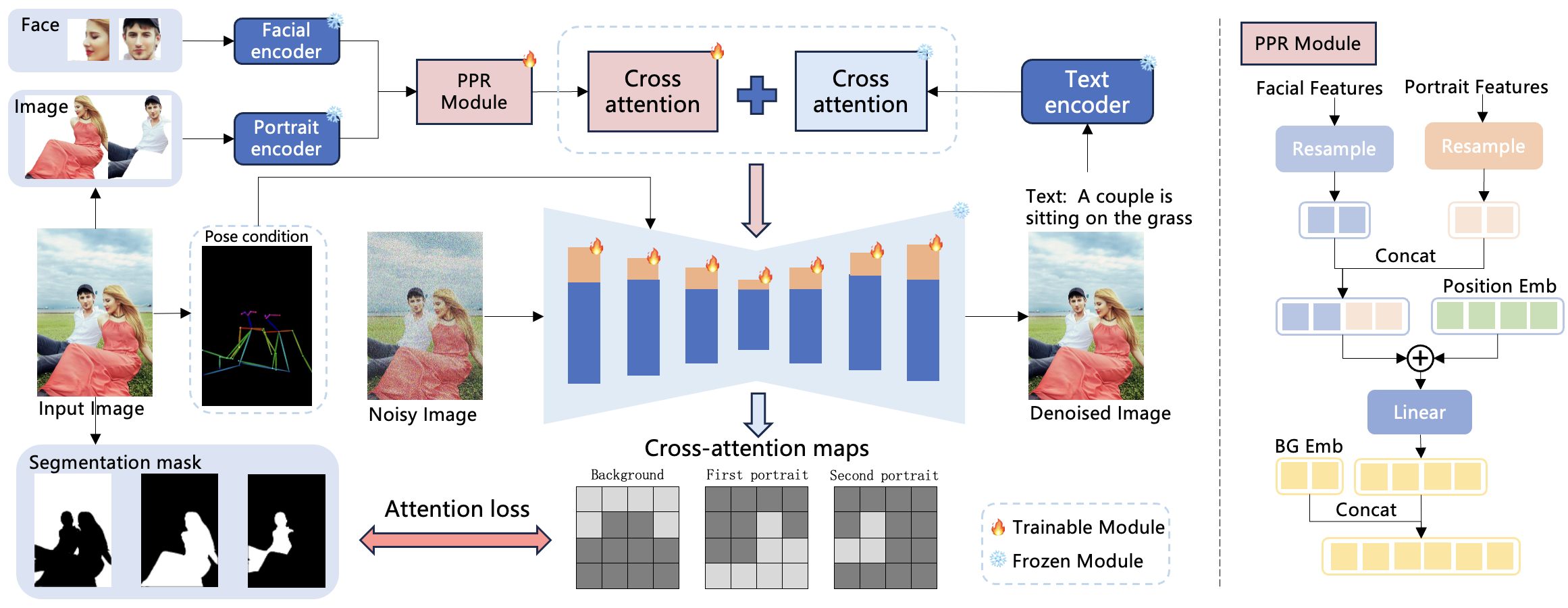}
	\caption{he model architecture of our proposed StoryMaker. The facial image and character image are embedded using the face encoder and image encoder, respectively, and refined through our proposed Positional-aware Perceiver Resampler module. Decoupled cross-attention with LoRAs is employed to inject these embeddings into the diffusion model. At the bottom, we illustrate the attention loss on cross-attention maps with the segmentation mask. The core of the PPR module is also depicted on the right.}
	\label{fig:pipeline}
\end{figure}

\section{Preliminaries}
We build our model on the state-of-the-art text-to-image model, namely Stable Diffusion XL \citep{podell2023sdxl}. In this section, we first introduce preliminaries about diffusion models and IP-adapter \citep{ye2023ip-adapter}, which form the foundation of our method.

\subsection{Stable diffusion}
The innovation of Stable Diffusion resides in executing the diffusion process within a low-dimensional latent space to enhance computational efficiency. This approach incorporates three primary components: a variational autoencoder (VAE) \citep{kingma2013auto} for compressing input images into the latent space, a text encoder to transform textual prompts into embeddings, and a U-Net \citep{ronneberger2015u} for the denoising procedure. For a given input image $x$ of dimensions $H\times W\times 3$, the VAE encoder $\varepsilon$ transforms it to a latent representation $z_0=\varepsilon(x)$ of dimensions $H/8\times W/8\times C$, where $8$ is the downsampling factor and $C$ is the latent dimension. The denoising process employs a U-Net $\epsilon_{\theta}$ to denoise the normally-distributed noise $\epsilon$ added to the noisy latent $z_t$ at timestep $t$, conditioned on $c$. Here, $c$ denotes the text embeddings generated by the pre-trained CLIP text encoder. The overall training objective is defined as:
\begin{center}
\begin{equation}
\mathcal{L}_{SD}=\mathbb{E}_{\varepsilon(x),c,\epsilon\sim\mathcal{N}(0,1),t}\left[\left\|\epsilon-\epsilon_{\theta}(z_t,t,c)\right\|_2^2\right]
\end{equation}
\end{center}
During inference, a random noise $z_t$ is drawn from Gaussian noise and iteratively denoised by the U-Net to yeild the initial latent representation $\hat{z_0}$. Subsequently, the VAE
decoder $D$ converts the initial latent into the pixel space as $\hat{x} = D(\hat{z_0})$.

\subsection{IP-Adapter}
IP-Adapter \citep{ye2023ip-adapter} introduces an image prompt adapter that allows the diffusion model to generate images conditioned on an image prompt. The method comprises two components: an image encoder to extract features from the reference image, and an adapter module with decoupled cross-attention layers to integrate these image features into the pre-trained text-to-image model. Specifically, in the original cross-attention layer of the diffusion model, given the query features $Z$ and text features $c_t$, the output of cross-attention $Z_t$ is defined by the following equation:
\begin{equation}
Z_t = Attention(Q,K_t,V_t) = Softmax(\frac{QK_t^T}{\sqrt{d}})V_t,
\end{equation}
where $Q=ZW_q$, $K=c_tW_k^t$, $V=c_tW_v^t$ are the query, key, and value matrices for the attention operation, respectively, and $d$ denotes the channel dimension of the feature. The newly introduced decoupled cross-attention is calculated as follows:
\begin{center}
\begin{equation}
Z_{new} = Attention(Q,K_t,V_t) + \gamma \cdot Attention(Q,K_i,V_i),
\end{equation}
\end{center}
where $c_i$ is the image prompt embed and $K_i=c_iW_k^{i},V_i=c_iW_v^{i}$ constitute the added attention operation for the image cross-attention. Here only $W_k^{i}$ and $W_v^{i}$ are the trainable weights.

\section{Method}
\subsection{Overview}
Given a reference image containing one or two characters, StoryMaker seeks to generate a series of new images featuring the same characters, maintaining not only identical faces, $i.e.$, identities, but also their clothing, hairstyles, and bodies. A narrative can then be created by altering the background, the characters' poses, and the style according to the text prompts.

 Specifically, we first extract the facial information, $i.e.$, identities, of the characters using the face encoder, and the details of their clothing, hairstyles, and bodies via the character image encoder. We then refine this information using the proposed Positional-aware Perceiver Resampler. To control the backbone generation network, we inject the refined information into the decoupled cross-attention module proposed by IP-Adapter \citep{ye2023ip-adapter}. To prevent multiple characters and the background from interleaving, we constrain the impact region of the cross-attention for different characters and background separately. ID loss is additionally utilized to maintain the characters' identities. Furthermore, to decouple pose information from the reference image, we train the network conditioned on detected poses by ControlNet \citep{zhang2023adding}. For enhanced fidelity and quality, we also train the U-Net with LoRA \citep{hu2021lora}. Once trained, we can either discard the entire ControlNet and control the characters' poses through text prompts or guide image generation with new poses during inference. The complete pipeline of our proposed method is illustrated in Figure \ref{fig:pipeline}.

\subsection{Reference Information Extraction}
Since the facial features extracted by the face recognition model effectively capture semantic details and enhance fidelity, similar to InstantID \citep{wang2024instantid} and IP-Adapter-FaceID \citep{ye2023ip-adapter}, we utilize Arcface \citep{deng2019arcface} to detect faces and obtain aligned facial embeddings from the reference image. To maintain consistency in hairstyles, clothing, and bodies, we first segment the reference image to crop the characters. Following recent works such as IP-Adapter \citep{ye2023ip-adapter} and MM-Diff \citep{wei2024mm}, we use the pretrained CLIP vision encoder, known for its rich content and style, to extract features of the hairstyles, clothing, and bodies of the characters. During training, the face encoder, $i.e.$, Arcface model, and the image encoder, $i.e.$, CLIP vision encoder, are kept frozen.

\subsection{Reference Information Refinement by Positional-aware Perceiver Resampler}
Following InstantID \citep{wang2024instantid} and IP-adapter \citep{ye2023ip-adapter}, we utilize two independent resampler modules to transform the facial features, $i.e.$, $F_{face}$, and the character features, $i.e.$, $F_{character}$, into facial embeddings and character embeddings, respectively. These embeddings are concatenated and augmented with positional embeddings, i.e., $E_{pos}$, which serve to distinguish different characters. To differentiate the foreground from the background, we introduce a learnable background embedding, $i.e.$, $E_{bg}$ and concatenate it into the final embedding. Denoting the two independent resampler modules as $R_1$ and $R_2$, the Positional-aware Perceiver Resampler is formulated as follows:
\begin{align}
E_1 &= R_1(F_{face}) \\
E_2 &= R_2(F_{character}) \\ 
E_i &= MLP(Cat(E_1, E_2) + E_{pos}) \\
c_i &= Cat(E_{bg}, Reshape(E_i, (N\ast{L},D)) \label{eq:A} 
\end{align}
where $L$ represent the number of tokens and the dimension of the character embeddings, respectively, and $N$ denotes the number of characters in the reference image. The image prompt embed for image cross-attention is $c_i$. We denote the $L$ tokens of the background embedding as $E_{bg}$, resulting in the dimension of $c_i$ is $((N+1)\cdot L)\times D$.

\subsection{Decoupled Cross-attention}
After extracting the reference information, we utilize the decoupled cross-attention to embed it into the text-to-image model, following IP-Adapter \citep{ye2023ip-adapter}.

\subsection{Pose Decoupling from Character Images}
Pose diversity is essential for storytelling. Training conditioned solely on character images can lead to the network overfitting to the poses of the reference images, resulting in generated characters with identical poses. To facilitate decoupling poses from character images, we condition the training on poses using Pose-ControlNet \citep{zhang2023adding}. During inference, we can either discard ControlNet and employ text prompts to control the poses of generated characters or guide generation with a newly provided pose.

\subsection{Training with LoRA}
Furthermore, to enhance ID consistency, fidelity, and quality akin to IP-Adapter-FaceID, LoRA layers \citep{hu2021lora} are integrated into each attention layer of the diffusion model. Specifically, in each cross-attention layer, $Q$,$K_t$,$V_t$,$K_i$ and $V_i$ are modified as follows:
\begin{equation}
\begin{cases}
Q &= Z(W_q+\Delta{W_q}), \\
K_t &= c_t(W_k^t+\Delta{W_k^t}), \\
V_t &= c_t(W_v^t+\Delta{W_v^t}), \\
K_i &= c_i(W_k^i+\Delta{W_k^i}), \\
V_i &= c_i(W_v^i+\Delta{W_v^i}) \label{eq:lora} \\
\end{cases}
\end{equation}
We freeze the U-Net model, and only the $\Delta{W}$ is trainable.

\subsection{Loss Constraints on Cross-attention Maps with Masks}
To prevent multiple characters and the background from interleaving, we regularize the influence region of cross-attention using the embeddings of different characters and the background. Unlike MM-Diff \citep{wei2024mm}, which does not consider the background, we introduce a learnable background embedding to address it. We constrain the influence region by calculating the MSE loss between the softmax values of cross-attention and the segmentation masks predicted by a pre-trained network. This design, $i.e.$, introducing a learnable background embedding, encourages a better separation not only within the foreground characters but also between foreground and background. As seen in Equation\ref{eq:A}, the first $L$ tokens of image prompt $c_i$ represent the background, with each subsequent set of $L$ tokens representing each character. In each layer of image cross-attention, we obtain the cross-attention map $A$ of size $h\times w$ for each character by summing all its $L$ tokens as:
 \begin{align}
P &= Softmax(QK^T/\sqrt{d}),\\
A &= \displaystyle \sum_{k=1}^L P_k  
 \end{align}
 Our proposed attention loss $\mathcal{L}_{attn}$ can be formulated as follows:
\begin{equation}
\mathcal{L}_{attn}=\frac{1}{N+1}\displaystyle \sum_{k=1}^{N+1}\left\|A_k-M_k\right\|_2^2, 
\end{equation}
where $N$ is number of characters in the reference image, and "+1" represents the background. 

\subsection{Overall Loss}
In training, we average $\mathcal{L}_{attn}$ across all $M$ layers and combine it with the diffusion loss as follows:
\begin{equation}
\mathcal{L} = \mathcal{L}_{SD} + \frac{\lambda}{M}\sum_{l=1}^{M}\mathcal{L}_{attn}
\end{equation}
where $\mathcal{L}$ is our final training objective and $\lambda$ is a weighting scalar. 

\begin{table}
    \caption{Quantitative comparisons on character-conditioned generation.The best results are in \textbf{bold}.}
    \centering
    \small
    \begin{tabular}{cccccc}
    \hline
        Method & Multi-person & Clothing, etc. & Face Sim. $\uparrow(\%)$ & CLIP-T $\uparrow(\%)$ & CLIP-I $\uparrow(\%)$ \\
        \hline
        MM-Diff \citep{wei2024mm} & \CheckmarkBold  & \XSolidBrush & 60.70 & 19.81 & \underline{78.46} \\
        
        PhotoMaker-V2 \citep{li2024photomaker} & \XSolidBrush  & \XSolidBrush & 61.83 & \textbf{23.71} & 75.31 \\
        
        IP-adapter-FaceID \citep{ye2023ip-adapter} & \XSolidBrush  & \XSolidBrush & 66.66 & 22.97 & 68.72 \\
       
        InstantID \citep{wang2024instantid} & \XSolidBrush  & \XSolidBrush & \textbf{73.90} & \underline{23.39} & 72.50 \\
        
        \textbf{StoryMaker(Ours)} & \CheckmarkBold  & \CheckmarkBold & \underline{67.36}  & 21.75 & \textbf{79.51} \\
        \hline
    \end{tabular}
    
    \label{table1}
\end{table}

\section{Experiments}
\subsection{Setup}
\subsubsection{Datasets}
We collect an internal character dataset consisting of a total of 500K images, including 300K single-character images and 200K two-character images. Image captions are automatically generated using CogVLM \citep{wang2023cogvlm}. We employ the buffalo\_l \citep{deng2019arcface} model to detect and obtain the ID-embedding of each face. Character segmentation masks are acquired using our internal instance segmentation model.

\begin{figure}
	\centering
        \includegraphics[scale=0.4]{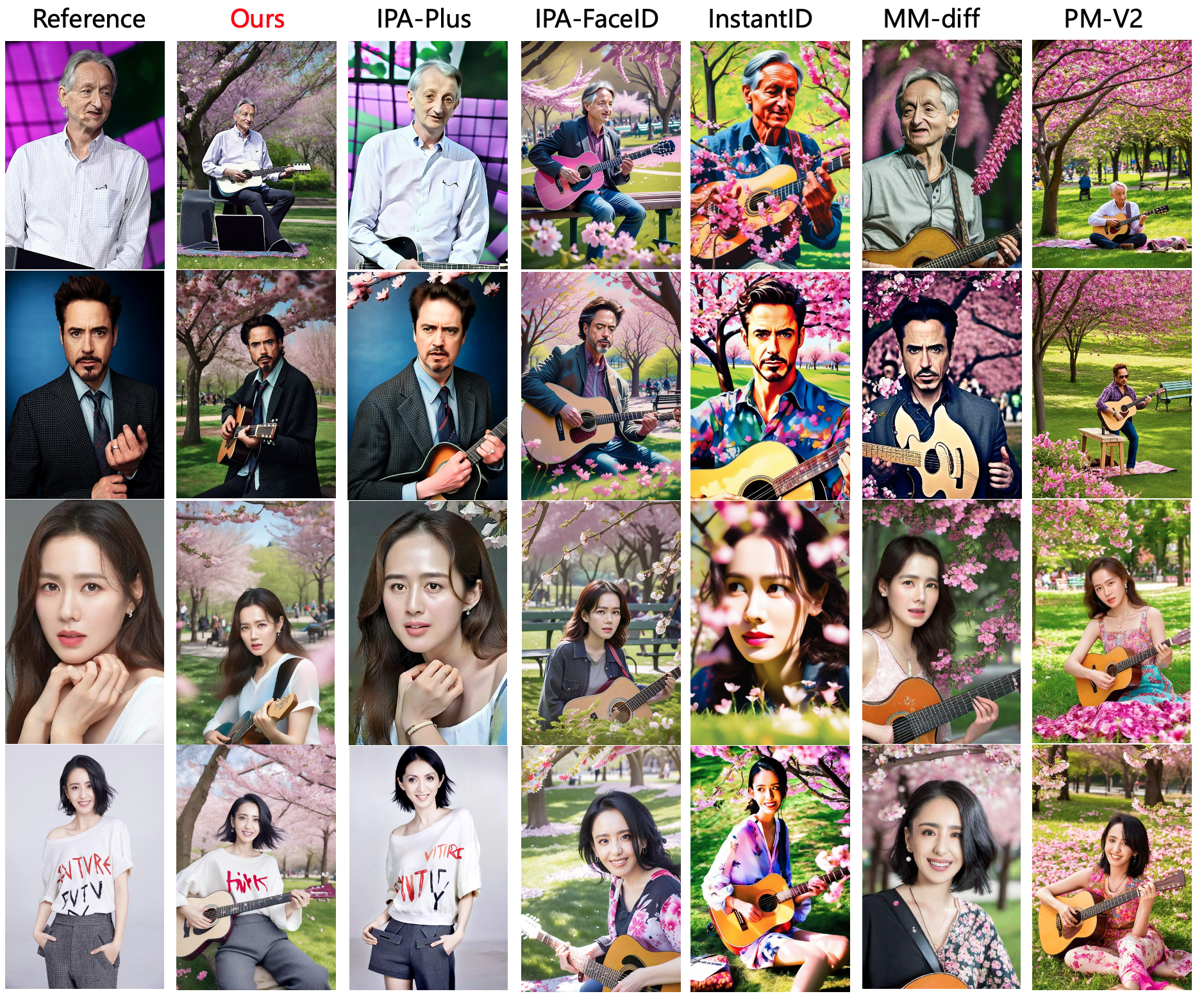}
	\caption{Visual comparison on single character condition generation.}
	\label{fig:compare}
\end{figure}

\subsubsection{Training Details}
We train our model based on Stable Diffusion XL \citep{rombach2022high}. Similar to IP-Adapter-FaceID \citep{ye2023ip-adapter}, we utilize buffalo\_l \citep{deng2019arcface} as the face recognition model and OpenCLIP ViT-H/14 \citep{ilharco_gabriel_2021_5143773} as the image encoder. The rank of trainable LoRA weights is set to 128. During training, we freeze the original weights of the base model and train only the PPR module and LoRA weights. Additionally, we initialize the weights of the resample module for the face and character from IP-Adapter-FaceID and IP-Adapter, respectively. Our model is trained for 8k steps on 8 NVIDIA A100 GPUs with a batch size of 8 per GPU. We use AdamW with a learning rate of 1e-4 for the first 4k steps and 5e-5 for the last 4k steps. We set $\lambda$ to 0.1. Training images are resized to a 1024$\times$1024 resolution. The text caption is randomly dropped by 10\% during training, and the cropped character image is randomly dropped by 5\%. During inference, we use the UniPC \citep{zhao2024unipc} sampler with 25 steps and set the classifier-free guidance to 7.5.

\subsubsection{Evaluation Metrics}
To compare with other methods, we evaluate our methods in a single-character setting. We collect a dataset of 40 characters and adopt 20 unique text prompts from FastComposer \citep{xiao2023fastcomposer} and generate 4 images for each prompt. Following FastComposer \citep{xiao2023fastcomposer} and MM-Diff \citep{wei2024mm}, we use CLIP image similarity (CLIP-I) to compare the generated images with reference images. For identity preservation, we employ buffalo\_l \citep{deng2019arcface} model to detect and calculate the cosine similarity (Face Sim.) between two face images. Additionally, we assess the image-text similarity using the CLIP-score (CLIP-T).

\subsection{Results}
\subsubsection{Quantitative Evaluation}
As shown in Table \ref{table1}, we compare our StoryMaker with four tuning-free character generation models, including MM-Diff \citep{wei2024mm}, PhotoMaker-V2 \citep{li2024photomaker}, InstantID \citep{wang2024instantid}, and IP-Adapter-FaceID \citep{ye2023ip-adapter}. Our proposed StoryMaker achieves the highest CLIP-I score among previous methods due to the consistency of the entire portrait, including face, hairstyle, and clothing, though it has a relatively lower CLIP-T, slightly compromising text prompt adherence. For face similarity, our method outperforms others except for InstantID. We attribute InstantID's superior performance to the extensive training data and the IdentityNet controlling module. It should be noted that among all evaluated methods, only MM-Diff and our method can preserve the ID of multiple persons. Moreover, StoryMaker is the only approach that maintains consistency not only in faces but also in clothing, hairstyles, and bodies.

\begin{figure}
	\centering
        \includegraphics[scale=0.38]{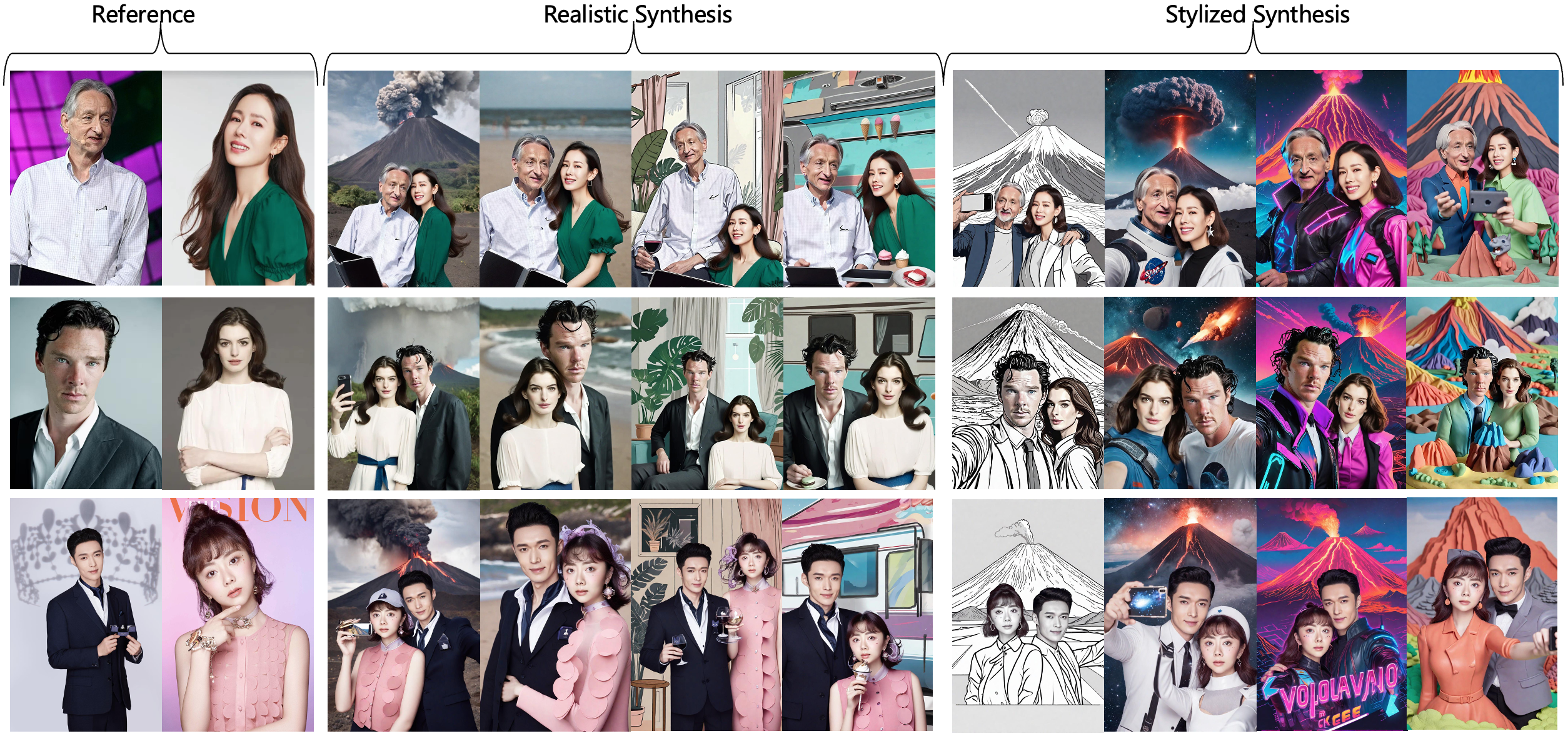}
	\caption{Visualization of two-character image generation. The first two columns display two different reference character images. The middle four columns illustrate StoryMaker's ability for realistic synthesis. The last four columns demonstrate results of stylized synthesis, where the character embedding is set to zero.}
	\label{fig:two}
\end{figure}

\subsubsection{Visualization}
\textbf{Single-Character Image Generation.} As shown in Figure \ref{fig:compare}, compared to IP-Adapter-FaceID, InstantID, MM-Diff, and PhotoMaker-V2, which are designed for identity preservation, the proposed StoryMaker not only maintains face fidelity but also clothing consistency. While IP-Adapter-Plus performs well on clothing consistency, it falls short in text prompts following and face fidelity.

\textbf{Multiple-characters Image Generation.} We further demonstrate the performance of multiple-character image generation. As shown in Figure \ref{fig:two}, with a text prompt, our method can generate different poses of two characters while maintaining consistency in faces, clothing, and hairstyles. Additionally, due to the use of two independent resampler modules, we can set the character embedding ($E_2$ in Equation \ref{eq:A}) to all zero, while maintaining only ID-preserving and generating stylized synthesis in the last four columns in Figure \ref{fig:two}.

\textbf{Personalized Story Diffusion.} Given reference character images, our proposed StoryMaker can generate consistent character images based on arbitrary prompts, enabling the creation of a story using a series of prompts. As illustrated in the top three rows of Figure \ref{fig:day1}, our method generates a series of images of a single person according to a short story composed of five text prompts describing "A day in the life of an office worker." The poses of the generated characters vary without being controlled by given pose maps. In the bottom two images of Figure \ref{fig:day1}, we present a story featuring the movie "Before Sunrise," generated with two characters. To achieve optimal results, we control the generation using specified poses. 

\begin{figure}
	\centering
        \includegraphics[scale=0.43]{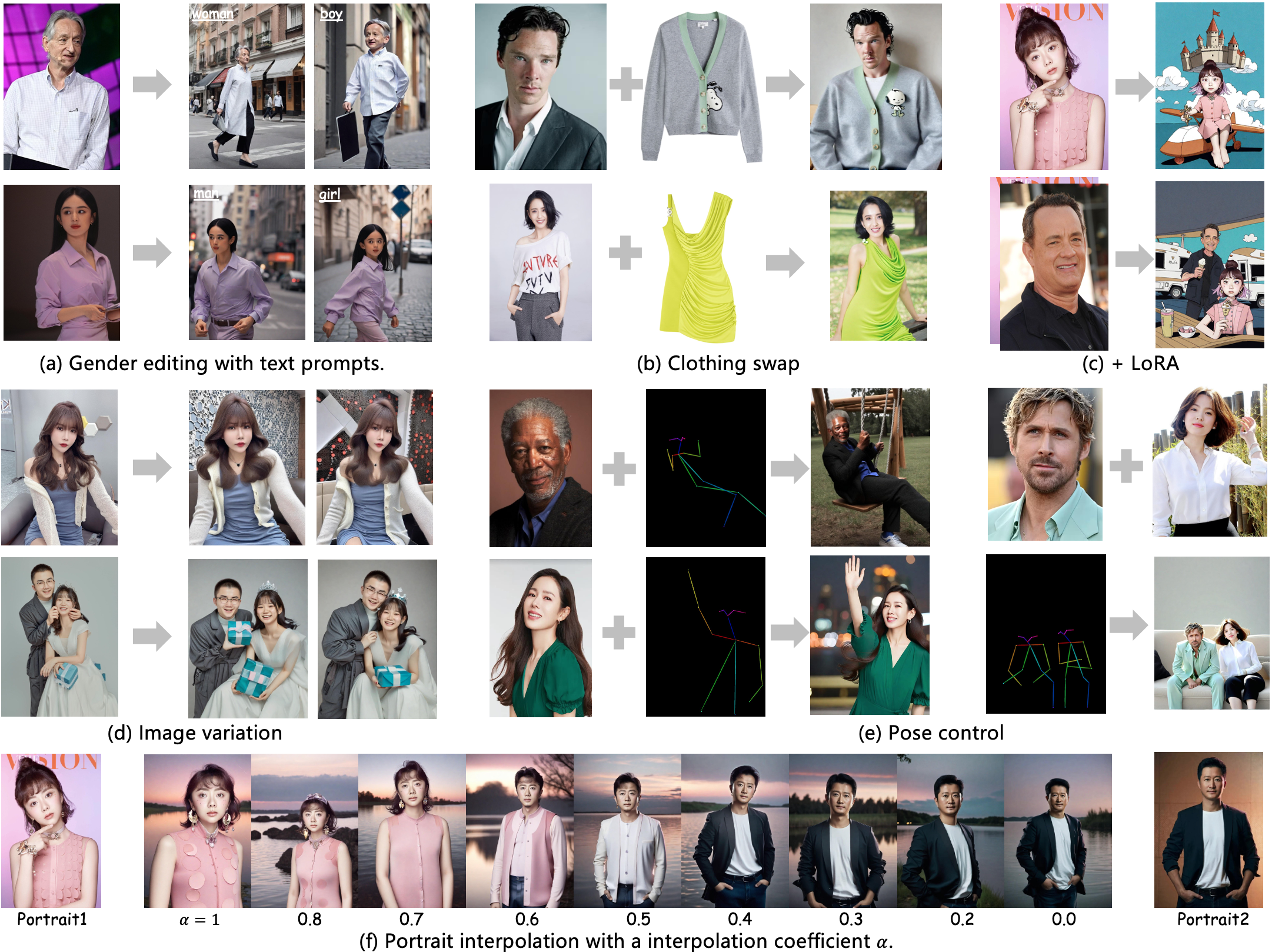}
	\caption{Diverse applications of StoryMaker.}
	\label{fig:diverse}
\end{figure}

\textbf{Applications.} The excellent performance of our method in aligning IDs, clothing, maintaining prompt consistency, and enhancing the diversity and quality of generated images provides a strong foundation for diverse downstream applications. As shown in Figure \ref{fig:diverse}(a), a man or woman could become a boy or girl while maintaining clothing consistency. Additionally, StoryMaker demonstrates a surprising ability for clothing swapping (Figure \ref{fig:diverse}(b)), achieved by replacing the character image with a clothing image, indicating that the character embedding contains clothing information. Moreover, similar to IP-Adapter \citep{ye2023ip-adapter} and InstantID \citep{wang2024instantid}, StoryMaker functions as a plug-and-play module, capable of integrating with LoRA or ControlNet to generate diverse images while maintaining character consistency, as shown in Figure \ref{fig:diverse}(c,e). Due to the character-preserving capability, human image variations can be realized, as illustrated in Figure \ref{fig:diverse}(d). Furthermore, we explore character interpolation between two characters, showcasing StoryMaker's ability to blend features from multiple characters, as demonstrated in Figure \ref{fig:diverse}(f).

\section{Conclusion}
In this paper, we introduce StoryMaker, a novel approach for personalized image generation that excels maintaining consistency not only in facial identities but also in clothing, hairstyles, and bodies across multiple characters scenes. Our method enhances narrative creation by allowing background, pose, and style variations via text prompts, enabling diverse and coherent storytelling.
StoryMaker leverages the Positional-aware Perceiver Resampler to obtain distinct character embeddings by fusing the features extracted from the face image and the cropped character image. To prevent intermingling of multiple characters and the background, we separately constrain the cross-attention impact regions of different characters and the background using MSE loss with segmentation masks. By incorporating pose decoupling through ControlNet and fidelity enhancements with LoRA, StoryMaker consistently generates high-quality images with matched identities and visual consistency.
Our extensive experiments demonstrate StoryMaker's superior performance in maintaining character identity and consistency, especially in multi-character scenarios, outperforming existing tuning-free models. The model's versatility is further highlighted through various applications such as clothing swapping, character interpolation, and integration with other generative plug-ins.
We believe StoryMaker significantly contributes to personalized image generation and opens possibilities for wide applications in digital storytelling, comics, and beyond, where individuality and narrative coherence are essential.

\section{Limitations}
In the absence of an explicit pose guide, the posture of generated characters often exhibits anomalies and lacks harmony. Moreover, generating three or more characters simultaneously presents significant challenges. The fidelity and detail of the generated clothing remain unsatisfactory.

\bibliographystyle{unsrtnat}
\bibliography{references}  






\end{document}